\documentclass{article}

\usepackage[final]{neurips_2023}

\usepackage[utf8]{inputenc} 
\usepackage[T1]{fontenc}    
\usepackage{hyperref}       
\usepackage{url}            
\usepackage{booktabs}       
\usepackage{amsfonts}       
\usepackage{nicefrac}       
\usepackage{microtype}      
\usepackage{xcolor}         

\usepackage{comment}
\usepackage{graphicx}
\usepackage{subcaption}
\usepackage{amsmath}

\title{Quick and Accurate Affordance Learning}

\author{%
  Fedor Scholz \\
  Computer Science, Cognitive Modeling \\
  University of Tübingen \\
  \texttt{fedor.scholz@uni-tuebingen.de} \\
  \And
  Erik Ayari \\
  Computer Science, Cognitive Modeling \\
  University of Tübingen
  \AND
  Johannes Bertram \\
  Computer Science, Cognitive Modeling \\
  University of Tübingen
  \And
  Martin V.~Butz \\
  Computer Science, Cognitive Modeling \\
  University of Tübingen
}

\begin{document}

\maketitle

\begin{abstract}
Infants learn actively in their environments, shaping their own learning curricula.
They learn about their environments' affordances, that is, how local circumstances determine how their behavior can affect the environment.
Here we model this type of behavior by means of a deep learning architecture.
The architecture mediates between global cognitive map exploration and local affordance learning.
Inference processes actively move the simulated agent towards regions where they expect affordance-related knowledge gain.
We contrast three measures of uncertainty to guide this exploration: predicted uncertainty of a model, standard deviation between the means of several models (SD), and the Jensen-Shannon Divergence (JSD) between several models.
We show that the first measure gets fooled by aleatoric uncertainty inherent in the environment, while the two other measures focus learning on epistemic uncertainty.
JSD exhibits the most balanced exploration strategy.
From a computational perspective, our model suggests three key ingredients for coordinating the active generation of learning curricula:
(1) Navigation behavior needs to be coordinated with local motor behavior for enabling active affordance learning.
(2) Affordances need to be encoded locally for acquiring generalized knowledge.
(3) Effective active affordance learning mechanisms should use density comparison techniques for estimating expected knowledge gain.
Future work may seek collaborations with developmental psychology to model active play in children in more realistic scenarios.
\end{abstract}

\section{Introduction}

Humans learn internal models of their environment in order to interact with it in a flexible, adaptive, and context-dependent manner \cite{Butz2016}.
Which models are suitable at a certain point in time depends on the current state of the environment:
In order to be able to prepare a cup of tea, both a cup and tea must be available and within reach \cite{Kuperberg2021}.
This observation is captured by the psychological concept of affordances as introduced by \citeauthor{Gibson1986}:
affordances encode which behaviors are possible in a given world state \cite{Gibson1986}.
While it may be said that, on the level of motor commands, any bodily action is executable at any point in time, its outcome clearly depends on the current context.
For example, in the absence of a cup and tea, `preparing a cup of tea' actions will at best result in pantomime.
Therefore, we define affordances---slightly more general than \citeauthor{Gibson1986}---as any factors in the environment that locally influence the outcome or success of an agent's actions, that is, that afford particular interactions and prohibit others. 

But what if an agent wants to learn how to prepare a cup of tea in a situation where neither is in reach?
In order to actively search for a location where critical preconditions are met, an allocentric map that relates coordinates to locally available affordances is needed.
Such a map enables an agent to search in allocentric space where to satisfy its personal motivations. 
While learning, it can use the map to actively explore affordances by maximizing expected information gain.
By focusing on aspects that can be learned and disregarding aspects that cannot be learned, we emulate curiosity and boredom, and thereby let the agent create its own learning curriculum \cite{Smith2018}.

Neuroscientific evidence suggests that the brain indeed learns such cognitive maps.
It was shown that such maps enable not only the navigation in allocentric, world-centered spaces but also the instantiation of local circumstances for reasoning and planning as well as for reflecting on the past and imagining potential futures \cite{Bottini:2020, Buckner:2007, Tolman:1948, O'Keefe:1978}.
To date it remains unclear, though, how such dual-use maps may be learned.
Here we assume the availability of an allocentric cognitive map for navigation and for providing sensory cues about local circumstances.
We fully focus on studying how to navigate the environment to effectively learn about the affordances the environment offers.


We showcase our reasoning in an artificial world with a simulated agent.
The agent perceives its environment via sensors and is able to imagine navigating it utilizing the provided cognitive map.
The environment it lives in is confined by borders and contains terrains that influence the behavioral dynamics of the agent in distinct manners: 
obstacles block the passage;  
force fields accelerate the agent in a certain direction;
fog fields corrupt the sensory signals with noise, mimicking an aleatoric uncertainty region.
Our study shows how navigation behavior may target affordance-respective knowledge gain, that  affordances should be encoded egocentrically, and that expected knowledge gain may best be computed by means of information-theoretic belief density comparisons.

\section{Background}

The problem setting we are concerned with can be described as a Markov Decision Process (MDP), where an agent receives observations from an environment and, based on that, executes actions that presumably lead to a desired state.
To this end, world models that encode visual information for planning were introduced before.
\citeauthor{Ha2018a} trained a vision model to produce codes that aid a controller in action selection \cite{Ha2018a}.
Since their vision model was trained as an autoencoder, though, it did not specifically produce codes that facilitate the controller's performance.
Therefore, we do not regard the outputs of their vision model as \emph{affordance} codes:
They do not necessarily extract behavior-relevant information, but are produced to reconstruct the visual input.
\citeauthor{Qi2020} went one step further by training a neural network to encode behavior-relevant information \cite{Qi2020}.
An agent was put into an environment to gather information about regions of harm and no harm.
The experiences were backpropagated onto the input of the visual system, producing affordance maps.
Subsequently, they trained a convolutional neural network to generate these maps from the visual input.
Therefore, the architecture was not trained in an end-to-end fashion.
As a result, the codes produced by the neural network were not optimized for behavioral control, which was anyway performed by a hard-coded A* algorithm.

In contrast to these studies, our work focuses on learning and exploration of a mapping from positions to affordances while a fixed, allocentric world model is provided, namely the world itself.
This is in line with \citealt{Epstein2015}, where the authors present an architecture that learns actual spatial affordances for navigation \cite{Epstein2015}.
Their space of affordances, however, is limited to three predefined affordances specifically designed for navigation.
Similarly, their action selection algorithm is based on handcrafted advisors and does not plan into the future.

\subsection{Affordance architecture}

\begin{figure*}
\centering
\includegraphics[width=.8 \linewidth]{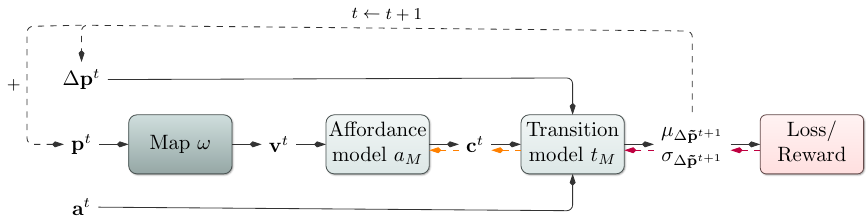}
\caption{
The overall architecture.
The look-up map $\omega$ provides visual representations $v_t$ of the environment at positions $p_t$.
The affordance model $a_M$ translates these representations into context codes $c_t$, which are utilized by the transition model $t_M$ for the generation of predictions in the form of expected positional changes $(\mu_{\Delta \tilde{p}^{t+1}}, \sigma_{\Delta \tilde{p}^{t+1}})$.
During training, the negative log-likelihood loss between predicted and observed $\Delta p^{t+1}$ observations is backpropagated to $t_M$ (red arrows) and further to $a_M$ (orange arrows), training both subcomponents end-to-end.
During control, potential behavioral interactions are evaluated via a reward function that combines estimates of epistemic knowledge gain with estimates of goal state proximity.
In this paper we fully focus on affordance learning and thus on epistemic knowledge gain. 
}
\label{fig:architecture}
\end{figure*}

The first end-to-end trained affordance architecture that produces codes that are explicitly optimized for behavioral control was introduced in \citealt{Scholz2022}.
In this case, the world model consists of a look-up map $\omega$, an affordance model $a_M$, and a transition model $t_M$ (see Figure \ref{fig:architecture}).
Given a position $\mathbf{p}_t$ in time step $t$, the hard-coded look-up map $\omega$ produces a sensory representation $\mathbf{v}_t$ of the environment at that position.
The affordance model is a convolutional neural network (CNN) that computes a context code $\mathbf{c}_t$ based on $\mathbf{v}_t$.
The transition model $t_M$---a multi-layer perceptron (MLP)---utilizes $\mathbf{c}_t$ as an additional input to predict the parameters of a probability distribution over positional changes $(\mu_{\Delta \tilde{\mathbf{p}}^{t+1}}, \sigma_{\Delta \tilde{\mathbf{p}}^{t+1}})$ given the last change in position $\Delta \mathbf{p}^t$ and the executed action $\mathbf{a}_t$.

Given sequences of position and action pairs, the model is trained end-to-end via backpropagation through time.
The loss is given by the negative log-likelihood of the observed change in position $\Delta 
\mathbf{p}^{t+1}$ in the predicted distribution.
This way, the affordance model $a_M$ tends to produce context codes $\mathbf{c}_t$ that facilitate accurate predictions by the transition model $t_M$.
The motor commands for data generation were selected randomly with a bias towards maintaining the same motor command for a few time steps.
We select a bias that ensures a comprehensive coverage of the environment, emulating an approximately uniformly distributed exploration with regards to positions.

Given a state and a policy, i.e., a sequence of actions, the world model enables the agent to imagine how the interaction with its environment unfolds over time.
In order to predict multiple time steps into the future, the predicted mean of the positional change is fed back into the model as the observed change in position.
The agent then probes the look-up map $\omega$ with the anticipated position, allowing it to probe the environment for local sensory representations.
We employ the cross-entropy method \cite{Rubinstein1999}, an evolutionary optimization algorithm, to infer the behavior that is expected to maximize reward and perform goal-directed control.

\subsection{Exploration}

The model was trained on previously generated sequences of observation and action pairs.
As mentioned before, a heuristic was used that led to sequences which covered the whole environment.
During the development of this heuristic it became apparent that it's exact implementation heavily influences the model's final performance and how fast it was able to learn.
Suboptimal heuristics lead to the agent getting stuck in corners, thereby neglecting other parts of the environment.

We conclude that it would be helpful for the agent to actively explore its environment based on estimates of potential information gain.
The agent should realize where its model is not able to generate accurate predictions and should thus explore those areas to improve its knowledge.
The affordance maps from above allow it to plan considering environmental circumstances, thus enabling it to focus on affordance learning.
To the best of our knowledge, such affordance-driven learning has not been explored before. 

Active exploration can be guided by uncertainty, because reducing uncertainty translates into a more accurate world model.
The mechanism should choose actions that produce high uncertainty in order to learn their effects in the current context and reduce uncertainty in the long run.
Usually, not all uncertainty can be reduced though.
As is often done in the machine learning community, we distinguish between two kinds of uncertainty: epistemic and aleatoric uncertainty \cite{Kiureghian2009, Hüllermeier2021}.
Epistemic uncertainty is inherent in the model.
It arises due to incompleteness or inaccuracy and can often be reduced by learning.
In contrast, aleatoric uncertainty is inherent in the environment and cannot be reduced by learning.
An example is the casting of a die, the outcome of which is practically unpredictable.
To learn affordances quickly, the exploration mechanism should disregard aleatoric uncertainty, as there is nothing to be learned from it.
Instead, it should choose actions that lead to high epistemic uncertainty.
In this way, the agent will choose actions that have the highest learning potential.
It is thus necessary to distinguish between aleatoric and epistemic uncertainties, to enable the active exploration of epistemic uncertainty while avoiding aleatoric uncertainty \cite{Vlastelica2021}.

\section{Methods}

The environment of our MDP is a 2-dimensional physics-based simulation.
The agent is represented by a circular, inert vehicle that is able to glide around by sending motor commands to its four rocket jets, which cause accelerations in four diagonal directions.
Observations consist of the last change in position $\Delta \mathbf{p}_t$ and a visual representation of the environment $\mathbf{v}_t$ at the currently considered position $\mathbf{p}_t$.
Actions determine to which extent the four jets are activated.
Their activities directly translate into accelerations.

In \citealt{Scholz2022}, the visual representation was given by a low-resolution image centered around the given position.
Due to the discrete nature of pixels, this approach introduced uncertainty:
the agent was not able to know exactly where, e.g., the borders of the environment were.
Therefore, in this work, the visual representation is given by distances to surrounding entities in eight directions.
Accordingly, we replace the CNN in the affordance model with an MLP.

All hyperparameters were optimized empirically, which led to the following configuration.
The affordance model $a_M$ is given by two linear layers with hidden sizes $64$ and $32$ and followed by ReLU activation functions.
Its output, representing affordance codes, is produced by a linear layer that maps onto size $5$ with the tanh activation function.
The transition model $t_M$ consists of a linear layer that maps onto size $64$, followed by a ReLU activation function, followed by two parallel linear layers, one for predicting $\mu_{\Delta \tilde{p}^{t+1}}$ without an activation function and one for predicting $\sigma_{\Delta \tilde{p}^{t+1}}$ with the exponential activation function.
We use Adam as our optimizer \cite{Kingma2014}.
For each experiment, we train $5$ different model instances based on different weight initializations.

\begin{figure}[htb!]
\begin{subfigure}[t]{0.49\linewidth}
    \includegraphics[width=\linewidth]{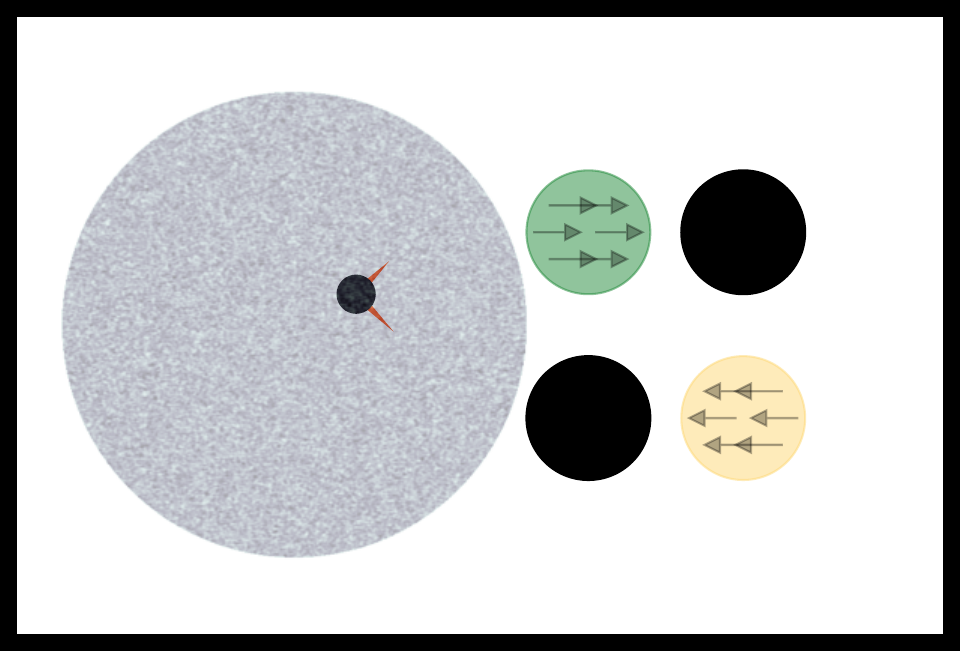}
    \caption{}
    \label{fig:env_train}
\end{subfigure}\hspace{\fill} 
\begin{subfigure}[t]{0.49\linewidth}
    \includegraphics[width=\linewidth]{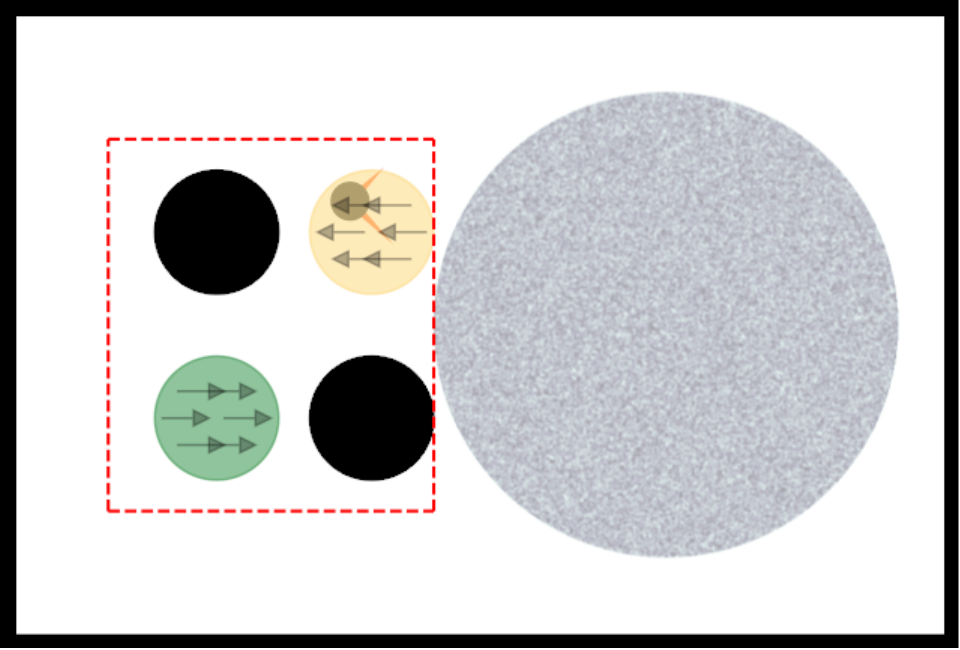}
    \caption{}
    \label{fig:env_val}
\end{subfigure}
\caption{
Environments used in our experiments.
A small circular agent (black) navigates its environment using four diagonally attached rocket jets (orange).
Fog fields are depicted in gray, obstacles in black. 
Force fields accelerate the agent to the right and left in green and yellow, respectively;
(a) depicts the environment used during training and (b) depicts the environment used during validation.
When focusing on affordance learning, we evaluate the model's performance only while the agent is within the red rectangle.
}
\label{fig:env}
\end{figure}

The environment is confined by borders and contains obstacles, both of which block the way.
Other terrains in the environment locally alter the sensorimotor dynamics of the vehicle.
Force fields accelerate the agent to the left or to the right and fog fields corrupt the observed position by Gaussian noise.
Borders, obstacles, and force fields produce affordances that can be learned by the agent, resulting in epistemic uncertainty until they are learned.
The uncertainty produced by fog fields cannot be reduced by learning and is therefore an instance of aleatoric uncertainty.
We use two different environments in our experiments, one for training and one for validation (see Figure \ref{fig:env}).

\subsection{Uncertainty estimation}
As shown in Figure \ref{fig:architecture}, a single model instance predicts the parameters of a normal distribution $P = \mathcal{N}(\mu, \sigma^2)$ over positional changes.
This way, the model performs uncertainty estimation.
However, it cannot distinguish between epistemic and aleatoric uncertainty.
In order to be able to efficiently guide exploration towards learnable interactions, the two types need to be disentangled.
We achieve this with ensembles of $i$ models that predict probability distributions $P_i$.
Since perfect models produce matching predictions, a measurement of an ensemble's members' disagreement can act as a proxy for epistemic uncertainty.
We investigate the following uncertainty measures:
\begin{itemize}
    \item Aleatoric uncertainty as the mean of the predicted standard deviations $\text{AU} = \mu(\mathbf{\sigma})$
    \item Epistemic uncertainty as the standard deviation of the predicted means $\text{EU}_\text{SD} = \sigma(\mathbf{\mu})$
    \item Epistemic uncertainty as the Jensen-Shannon divergence between predicted distributions $\text{EU}_\text{JSD} = \frac{1}{n} \sum_i D_\text{KL}(P_i || M),$ where $M = \frac{1}{n} \sum_i P_i$ and $D_\text{KL}$ denotes the Kullback-Leibler divergence
\end{itemize}
Using one of these uncertainty measures as the objective function during action selection allows the agent to perform exploration.
The difference between $\text{EU}_\text{SD}$ and $\text{EU}_\text{JSD}$ is that the former does not consider the disagreement in predicted standard deviations.
In contrast, JSD takes into account full distributions by extending the Kullback-Leibler divergence to potentially more than two distributions.
The JSD possesses additional advantageous properties, namely being bounded and symmetric for all distributions \cite{briet2008}.
Therefore, we anticipate $\text{EU}_\text{JSD}$ to provide the most precise measurements of epistemic uncertainty, leading to faster and more accurate affordance learning.

The training data is initialized with randomly generated sequences of observation and action pairs as in \citealt{Scholz2022}.
If an uncertainty-based exploration mechanism is employed, we gradually replace a subset of the training data after each epoch by new sequences.
These new sequences are generated by the agent itself via goal-directed control, based on behavior that is expected to maximize one of the above uncertainty measures.
Validation is performed on a dataset that is generated with the same heuristic as the training data is initialized with.

\section{Experiments}

We first investigate how globally vs locally informative sensory information results in different generalization capabilities.
Subsequently, we compare the different uncertainty measures with regards to their suitability for affordance learning.

\subsection{Globally vs locally informative sensory information}

The affordance model $a_M$ allows the agent to perceive its environment.
As input it receives distances to the nearest obstacle or terrain in each of eight directions.
In our initial experiment, we compare an agent's affordance learning capabilities with globally informative sensory information vs locally informative sensory information.
In the latter case, the distance sensors are limited in range, such that the agent is not able to perceive obstacles or terrains that are further away than a certain threshold.
We choose the threshold such that the agent is always able to perceive obstacles or terrains that could influence its dynamics in the next time step.
Globally informative sensory information is generated by sensors that are not limited in range.
Here, the perceived distances to the borders encode the current position of the agent in the environment.
For both bases, five model instances with differently initialized weights are trained on randomly generated sequences without any exploration taking place.

\subsubsection{Results}

\begin{figure}
\includegraphics[width=\linewidth]{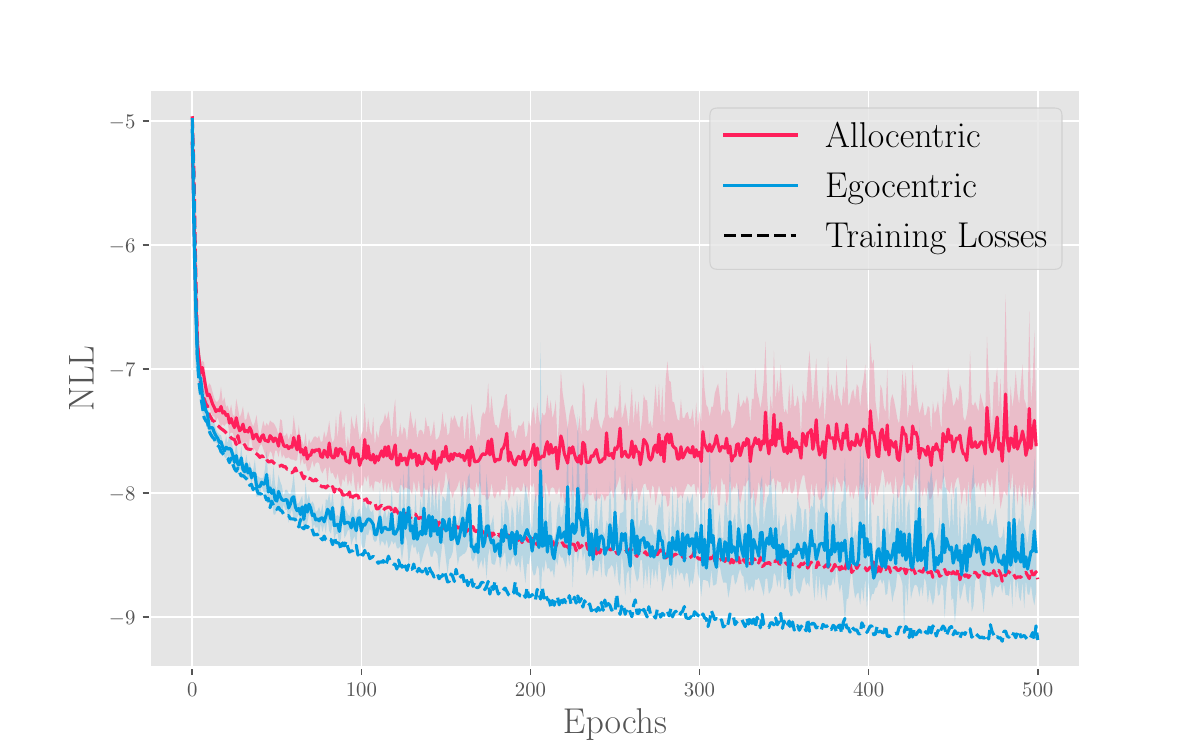}
\caption{
Losses for globally vs locally informative sensory information aggregated over $5$ seeds.
Shaded areas indicate standard deviations.
Solid lines represent validation losses, dashed lines show training losses.
The agent performs significantly better in the validation environment if equipped with distance sensors that are limited in range.
With distance sensors that are not limited, slight overfitting is observed.
}
\label{fig:results_exp1}
\end{figure}

We find that the model generalizes significantly better to the validation environment if equipped with locally informative sensory information (see Figure \ref{fig:results_exp1}).
Globally informative sensory information allows the agent to learn the environment ``by heart'', relating absolute coordinates to affordances, and thereby harming generalization capabilities:
The trained agent expects the obstacles and terrains to always be at certain absolute positions, an instance of overfitting.
With locally informative sensory information, however, the agent is able to learn only egocentrically encoded knowledge.
This kind of knowledge is applicable anywhere in the environment where the learned egocentric code applies.
This is the case in our validation environment, where the obstacles and terrains have the same effects on the agent but are positioned differently in the environment.

\begin{figure}
    \begin{subfigure}[t]{0.49\linewidth}
        \includegraphics[width=\linewidth]{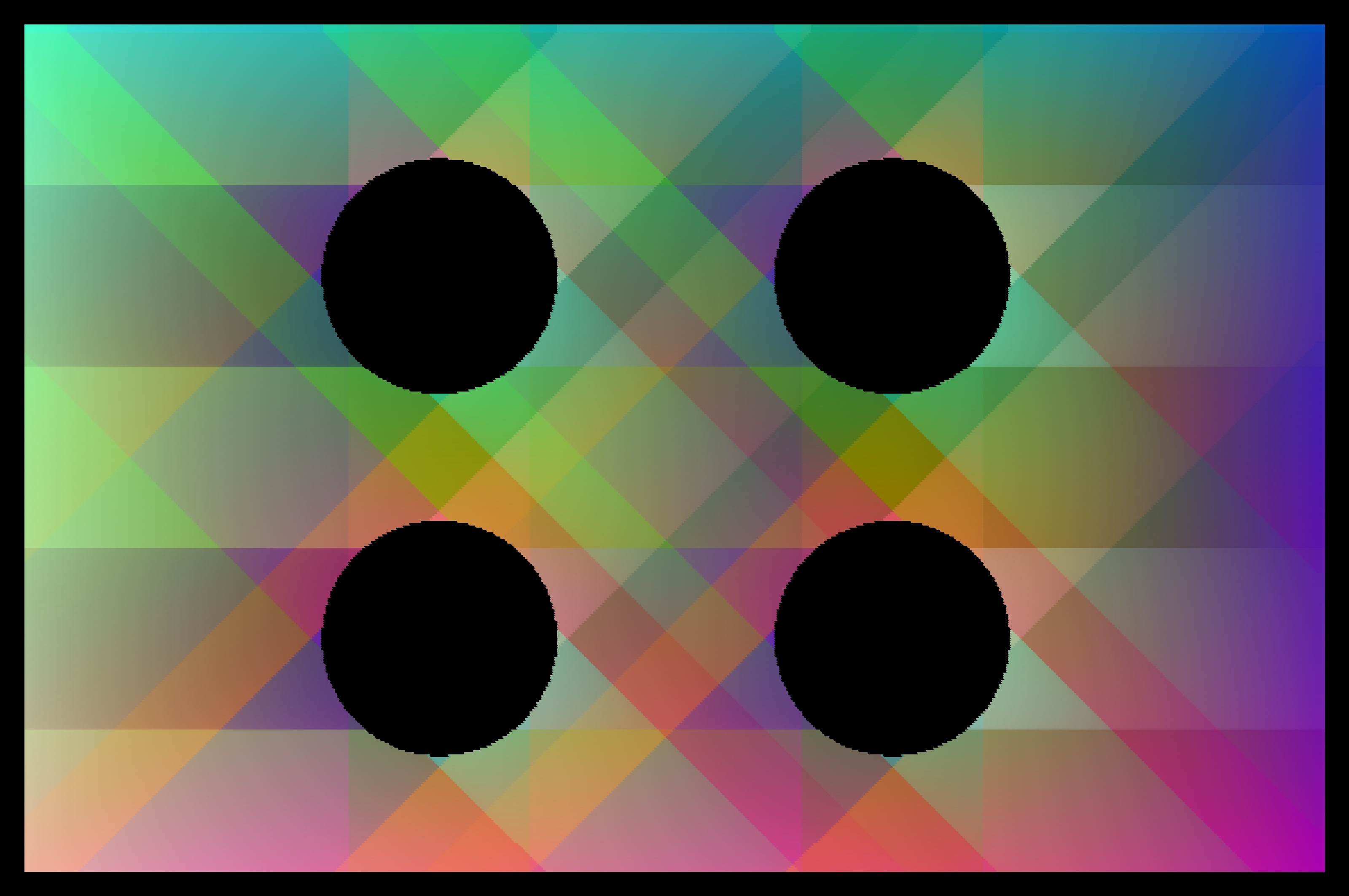}
        \caption{}
        \label{fig:affmap_exp1_global}
    \end{subfigure}
    \begin{subfigure}[t]{0.49\linewidth}
        \includegraphics[width=\linewidth]{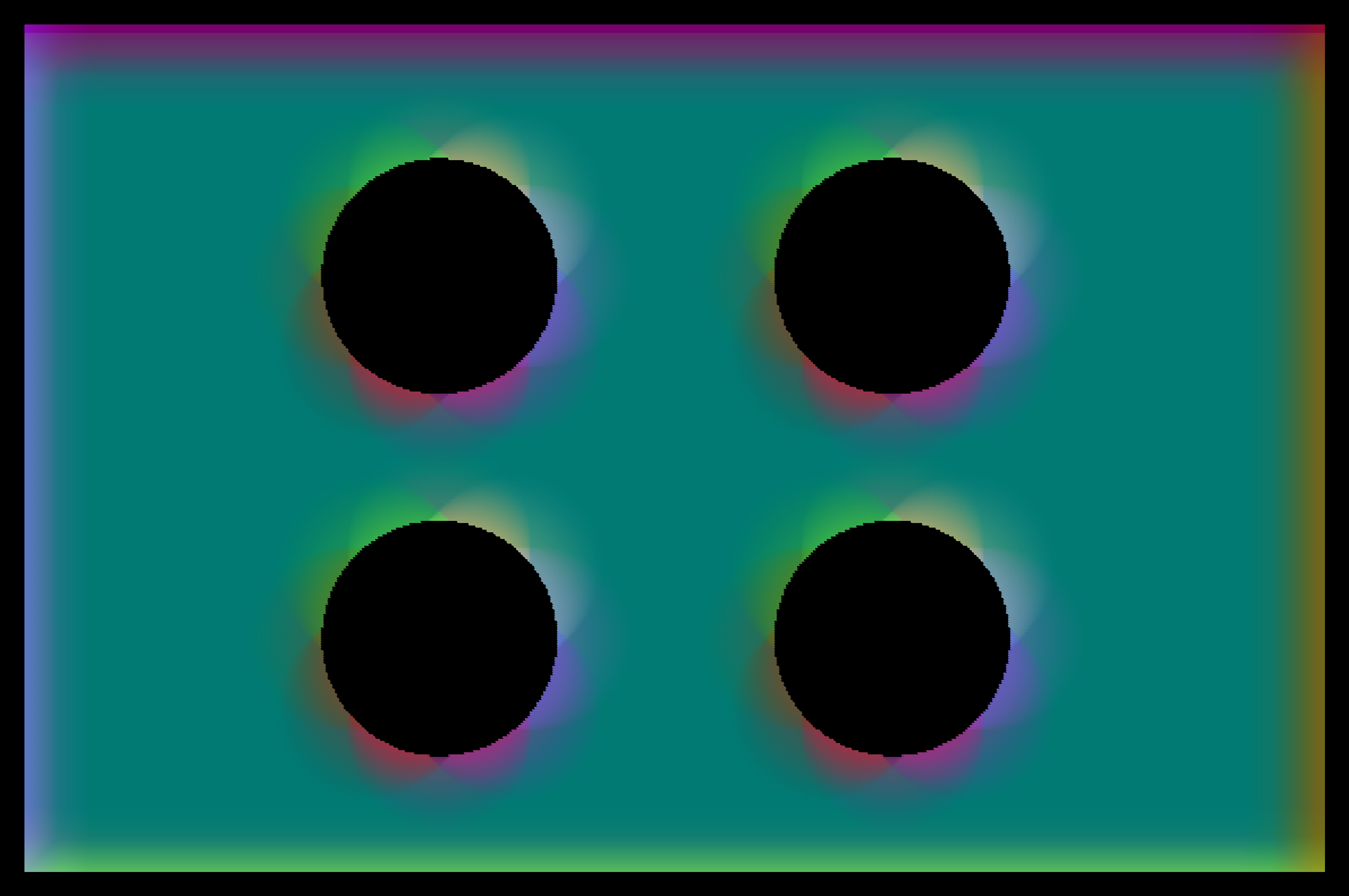}
        \caption{}
        \label{fig:affmap_exp1_local}
    \end{subfigure}
    \caption{
Affordance maps of an environment with four obstacles generated by a model with (a) globally informative sensory information vs (b) locally informative sensory information.
The maps are produced by feeding visual representations of the environment at regularly distributed positions into the affordance model and mapping the produced context code onto RGB space via principal component analysis.
True affordance maps, i.e., mappings from perceptual information to local behavioral constraints, only emerge in the latter case. Note how the obstacle's edges present the same local constraints as the borders, thus the matching borders show identical colors.} 
    \label{fig:affmaps_exp1}
\end{figure}

In order to gain an understanding of the model's inner workings, we visualize the produced affordance codes as maps.
To do so, we probe the environment at regularly distributed positions and feed the corresponding visual representations $\mathbf{v}_t$ into the affordance model.
A principal component analysis reduces the dimensionality from $5$ to $3$, allowing us to visualize the encoded affordances by RGB values.
These \emph{affordance maps} represent local behavioral possibilities, such as whether it is possible to move to the right.
Figure~\ref{fig:affmaps_exp1} shows the effects of the large-range distance sensors in a test environment with four obstacles:
While globally informative sensory information yields distorted affordance maps, the more local sensory signals indicate great generalization abilities. 

We have thus shown that it is rather advantageous to encode affordances in a local, egocentric manner. 
Therefore we use locally informative sensory information in the following experiments.

\subsection{Uncertainty-guided exploration}

We now investigate how affordances can be explored more efficiently, focusing on locally informative sensory information only.
Here, the agent replaces the $5\%$ oldest training sequences with new sequences that are generated by the exploration mechanism in each epoch.
We compare the different uncertainty measures $\text{AU}, \text{EU}_\text{SD}$, and $\text{EU}_\text{JSD}$.
The uncertainty measures are used as the objective that is to be maximized by the behavior inference mechanism.
We always compare performances to the \emph{random} behavioral policy, where the agent is not able to explore its environment in an active manner but is trained on a static training set.
We evaluate each condition using $5$ different seeds, with each seed generating an ensemble of size $5$.

\subsubsection{Results}

\begin{figure}
\includegraphics[width=\linewidth]{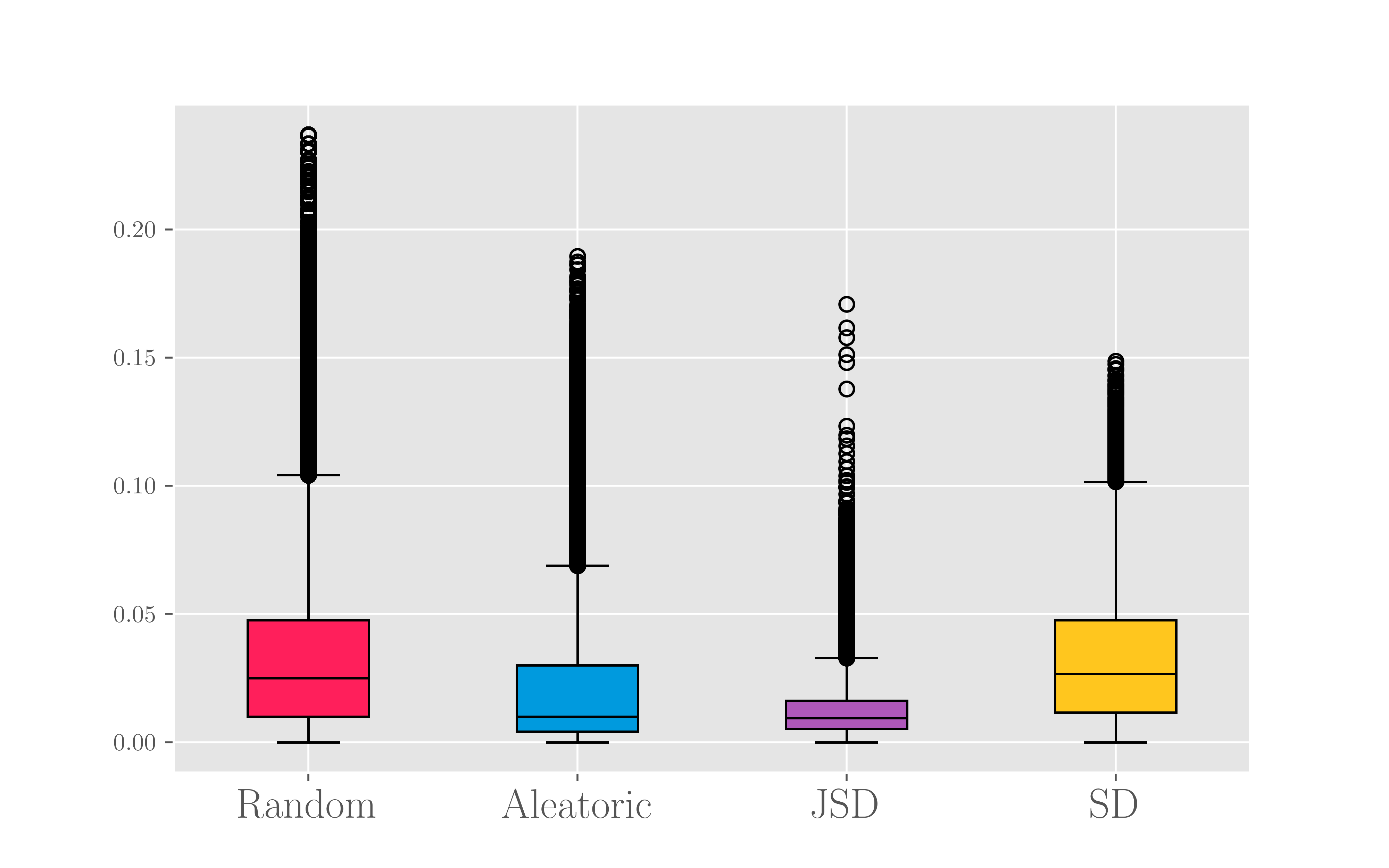}
\caption{
Boxplots of the velocities the agent exhibits during training with the different exploration mechanisms, taken over all epochs across the entire environment.
No uncertainty estimate produces velocities as high as the \emph{random} heuristic.
}
\label{fig:exp2_speeds}
\end{figure}

First, we examine the velocities the agent exhibits for the different cases defined as the distance traveled between two consecutive time steps (see Figure \ref{fig:exp2_speeds}).
We find that the agent produces significantly higher velocities if no active exploration takes place, i.e., in the \emph{random} condition.
This poses a disadvantage for the other cases as high velocities that would be present in the validation data are never encountered during training.
We therefore modify the validation set by adjusting the heuristic to produce lower velocities.
Further, we focus on learnable affordances rather than on areas with high sensory uncertainty by restricting the validation set to data points where the agent is within the red rectangle in Figure~\ref{fig:env}.

\begin{figure}
\includegraphics[width=\linewidth]{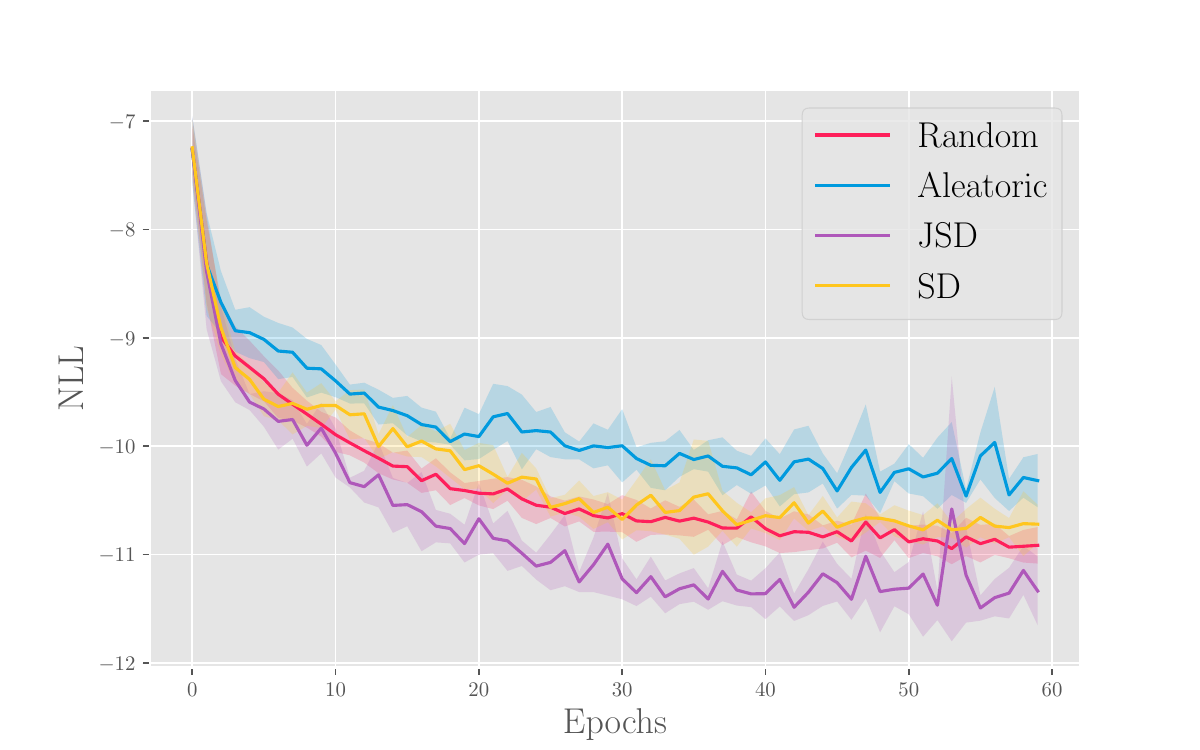}
\caption{
Losses for uncertainty-guided exploration aggregated over $5$ seeds in the adapted, affordance-focussed validation set.
Shaded areas indicate standard deviations.
The best performance is achieved if the agent explores its environment based on the $\text{EU}_\text{JSD}$ uncertainty measure.
It also allows the agent to learn affordances the quickest.
}
\label{fig:exp2_loss}
\end{figure}

The validation loss on the modified and restricted validation set is shown in Figure \ref{fig:exp2_loss}.
An agent that focuses on aleatoric uncertainty during exploration indeed performs worst when confronted with the learnable affordances.
$\text{EU}_\text{JSD}$-based behavioral inference learns the fastest and produces the lowest validation loss overall.

\begin{figure}
\begin{subfigure}[t]{0.49\linewidth}
    \includegraphics[width=\linewidth]{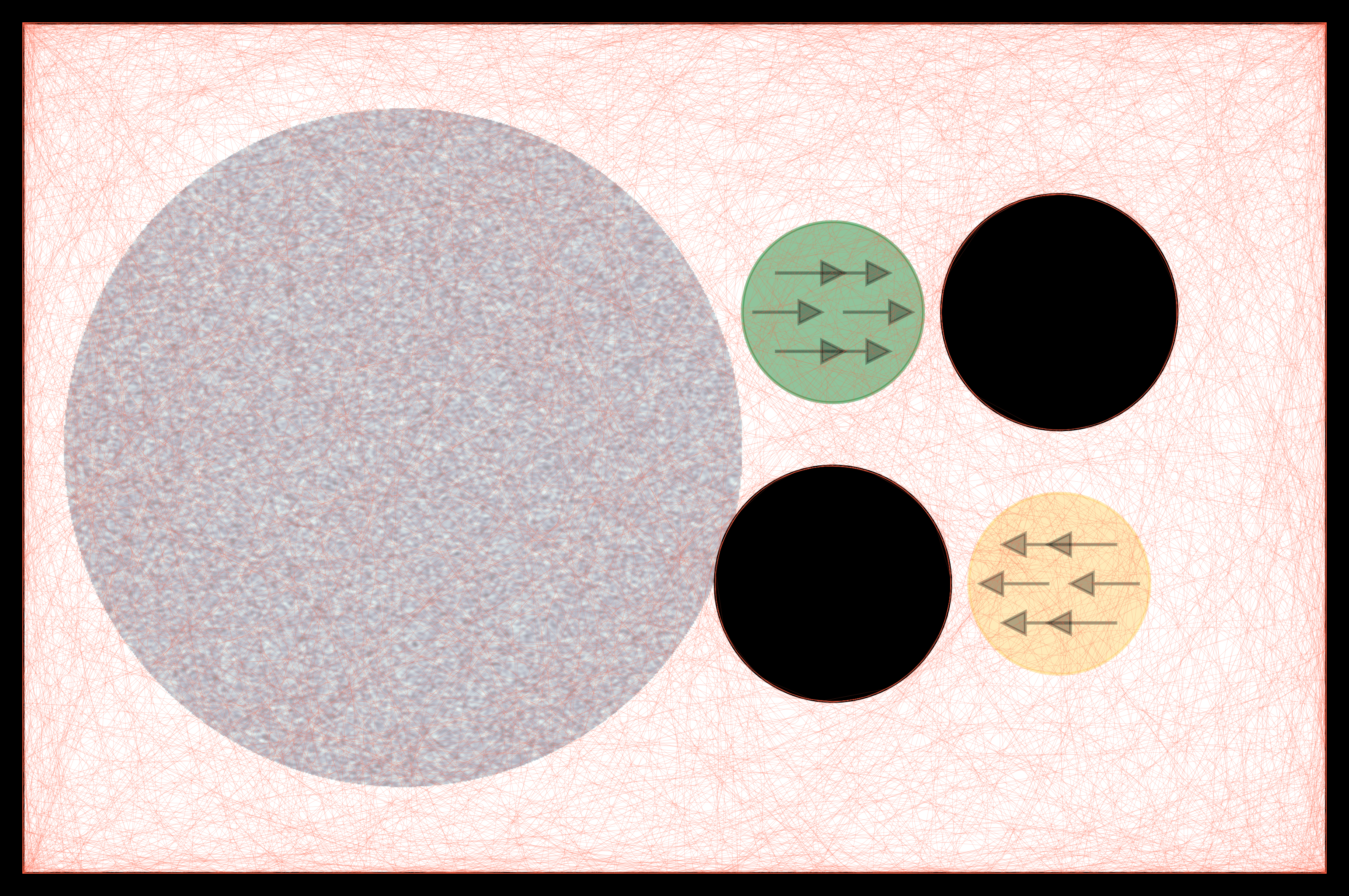}
    \caption{Random heuristic}
    \label{fig:exp2_heatmap_random}
\end{subfigure}\hspace{\fill} 
\begin{subfigure}[t]{0.49\linewidth}
    \includegraphics[width=\linewidth]{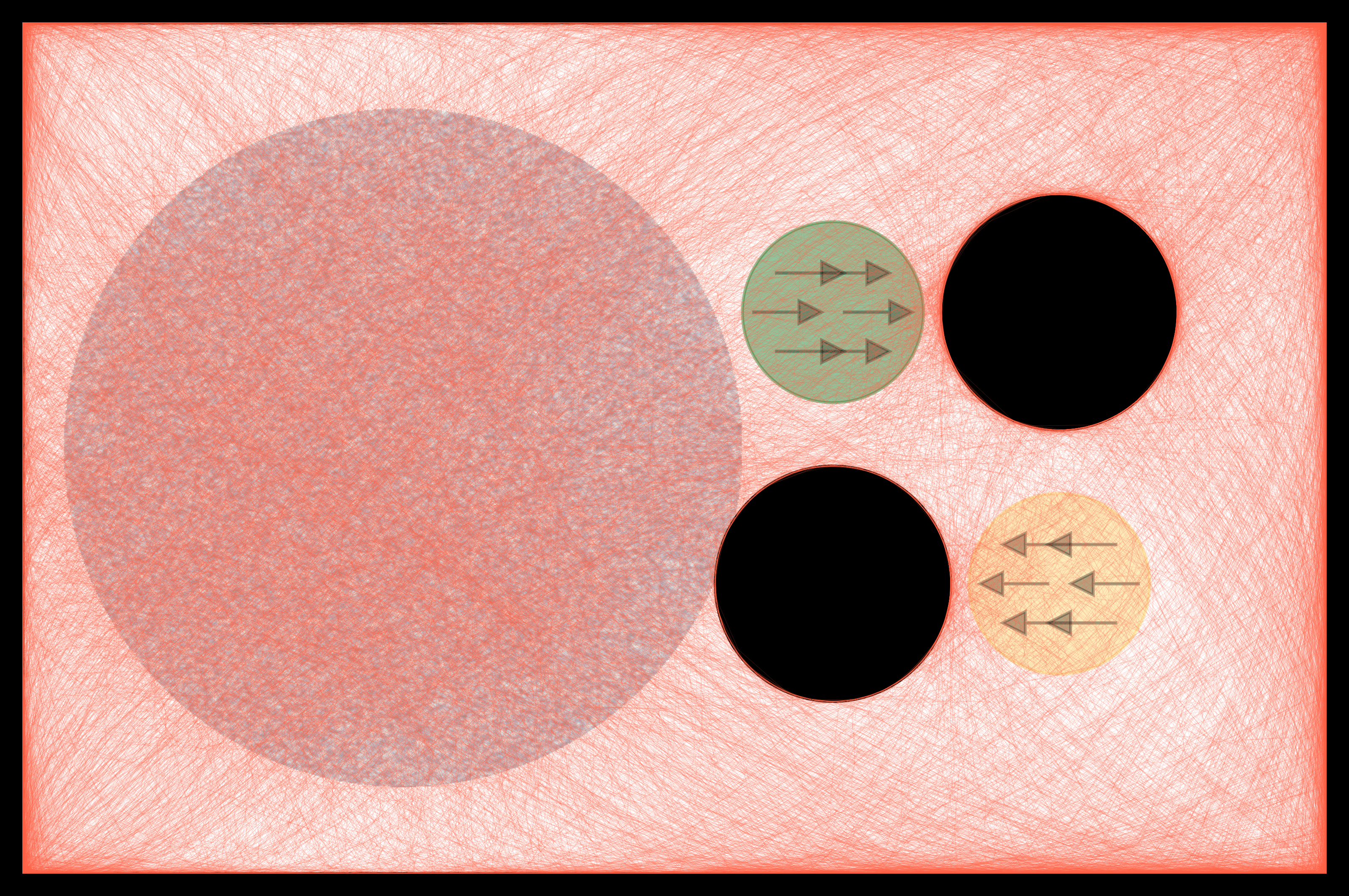}
    \caption{Aleatoric $\text{AU}$}
    \label{fig:exp2_heatmap_aleatoric}
\end{subfigure}
\bigskip 
\begin{subfigure}[t]{0.49\linewidth}
    \includegraphics[width=\linewidth]{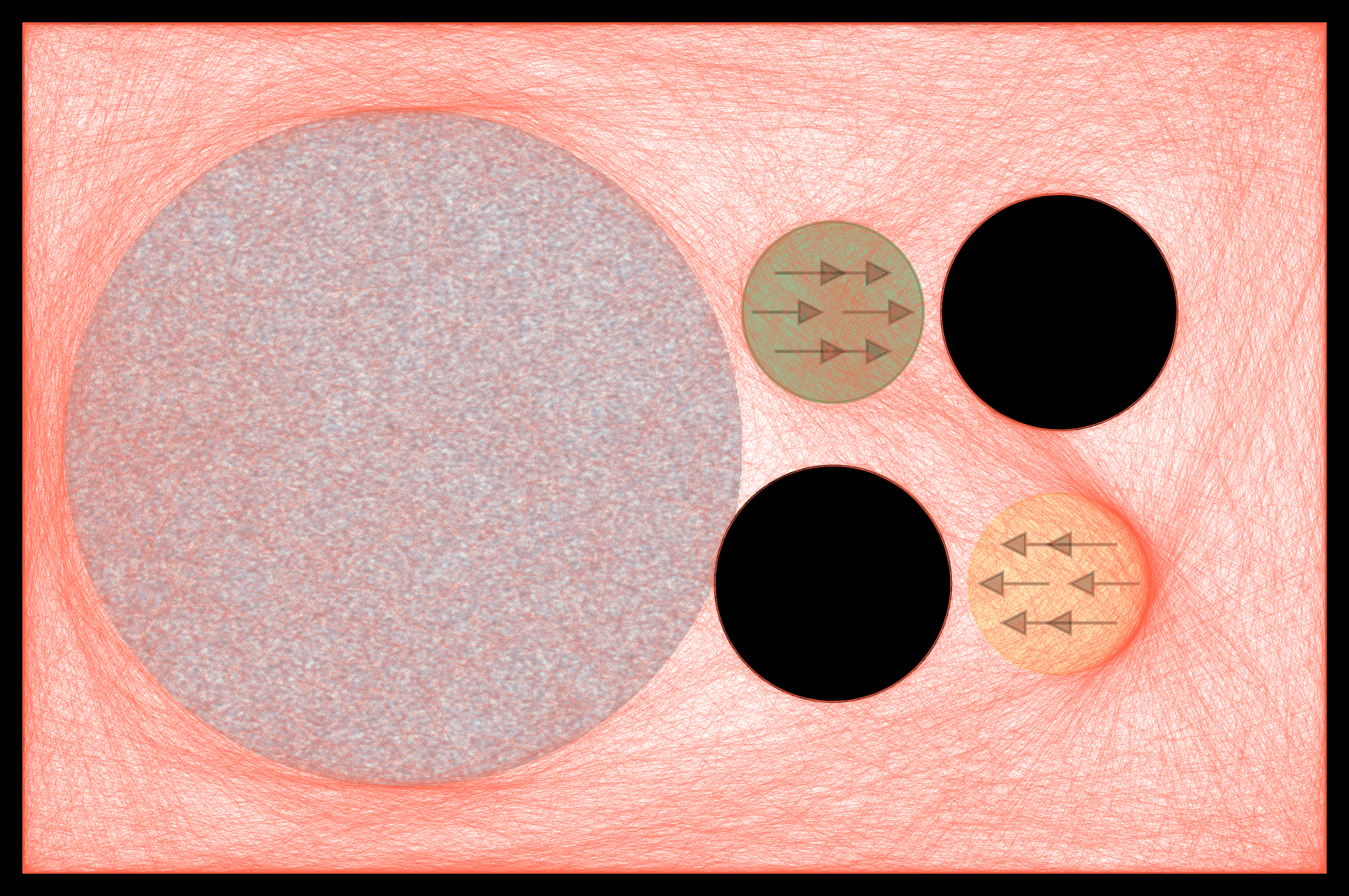}
    \caption{Epistemic $\text{EU}_\text{SD}$}
    \label{fig:exp2_heatmap_sd}
\end{subfigure}\hspace{\fill} 
\begin{subfigure}[t]{0.49\linewidth}
    \includegraphics[width=\linewidth]{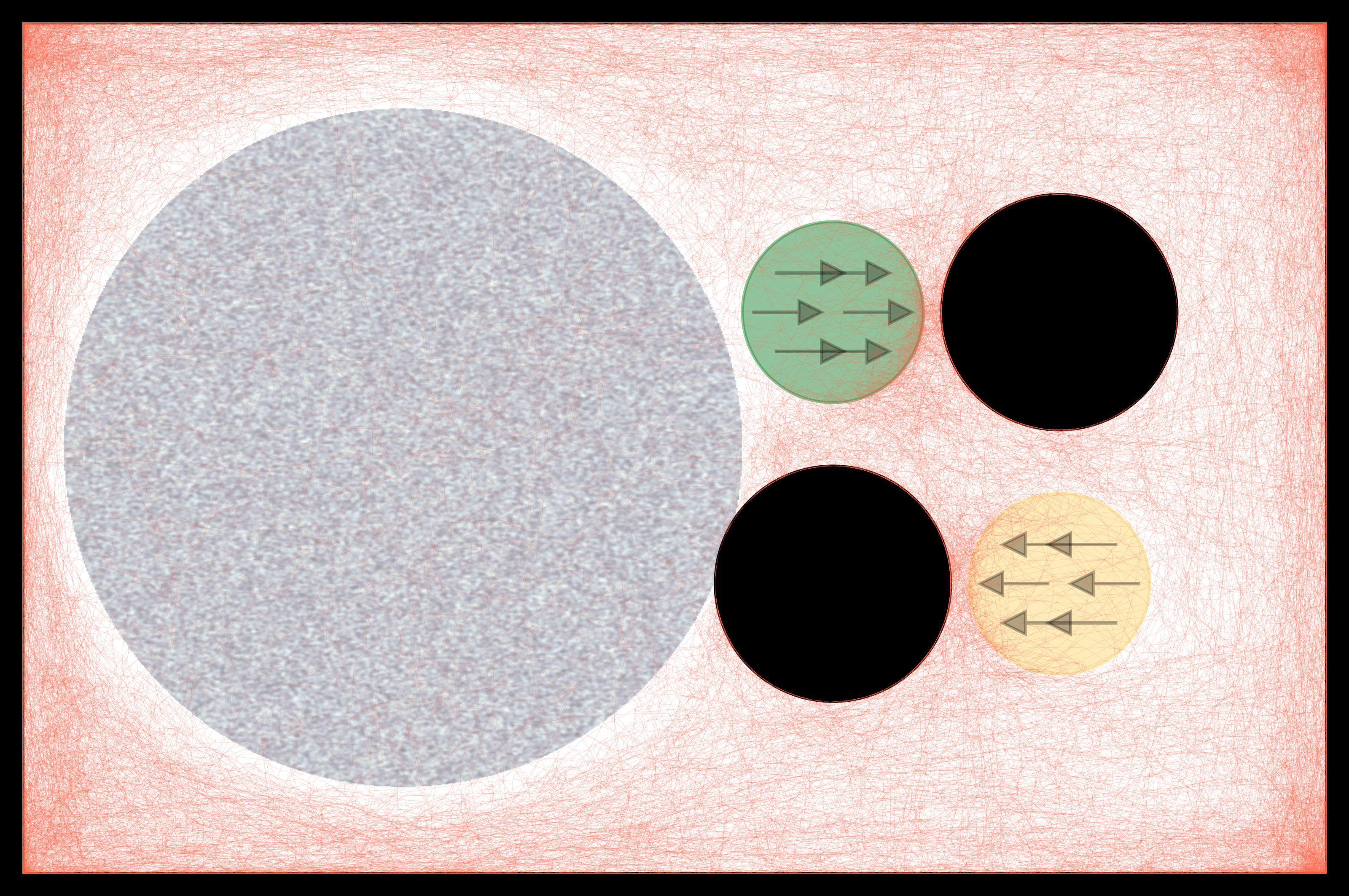}
    \caption{Epistemic $\text{EU}_\text{JSD}$}
    \label{fig:exp2_heatmap_jsd}
\end{subfigure}
\caption{
Positional heatmaps based on all sequences the agent sees in the corresponding conditions during training.
(a) The \emph{random} heuristic sees all locations in an approximately uniformly distributed manner. As the training set does not change over the course of training, the heatmap is less dense.
(b-d) When actively exploring the environment via the different uncertainty-based exploration strategies, the heat maps indicate most attractive sub-regions.
}
\label{fig:exp2_heatmaps}
\end{figure}

We also monitor the agent's positions and generate heatmaps to see which parts of the environment the agent explores the most with the different uncertainty estimates (see Figure \ref{fig:exp2_heatmaps}).
Training based on sequences generated by the random heuristic results in relatively uniform coverage of the environment by design.
We find that, of all approaches, exploration based on aleatoric uncertainty leads to the agent exploring the fog field the most.
With $\text{EU}_\text{SD}$, the agent is more focused on other affordances than the fog, partially avoiding regions of high aleatoric uncertainty.
The most profound effect, however, can be seen when $\text{EU}_\text{JSD}$ is employed: the agent avoids the fog field to the greatest extent while exploring the rest of the environment relatively uniformly.
We conclude that $\text{EU}_\text{JSD}$ is the most suitable uncertainty measures for quick and accurate affordance learning.


%
%
%
%

\section{Discussion}

This study investigated how an agent can learn affordances in a quick and accurate manner.
An artificial agent was put into a simulated environment and equipped with a cognitive map, which maps positions to sensory signals. The agent learned a predictive world model in the form of a neural network, predicting action consequences conditioned on local sensory information.
The agent explored its environment and learned about affordances that inform it about environmental aspects that locally influence its behavior.
Our results indicate that cognitive agents that actively learn about affordances should integrate three key ingredients in this process. 
First, search for novel affordances should be pursued within world-centered cognitive maps, which allow the activation of local views at particular positions within the map.
Second, the learning of affordances should focus on local, body-relative sensory encodings. 
Third, divergence measures between a small collection of model-predictive densities are best-suited to identify regions that support further model learning. 

We found that locality of perception is an important ingredient to allow for generalization, which is in accordance with the literature \cite{Epstein2015}.
If sensors are too globally informative, then local, generalizable affordances are hard to learn.
This is similar to infants which are born with low visual acuity \cite{Smith2018}, which may indeed be helpful to learn about global outlines and otherwise focus on local visual information, such as faces, hands, and objects.
Note that locality does not necessarily need to be in space, but could also be in time or respective other conceptual spaces \cite{Gaerdenfors:2014}.

The learned affordances can be depicted by relating positions in the environment to affordance codes.
The resulting affordance maps are related to cognitive maps and are possibly related to what the hippocampus is partially doing. 
Cognitive maps in the hippocampus appear to be essential for pursuing successful navigation and other environment-centered tasks. 
Meanwhile, the strong interconnectivity with neocortical areas indicates that particular local codes in the hippocampus may trigger views onto the local surrounding \cite{Mallot:2014, Roehrich:2014}.
Moreover, the mapping between allocentric positions and egocentric perceptions is probably mediated between the hippocampal loop and the rest of the default mode network \cite{Bottini:2020, Buckner:2008, Zacks:2020, Stawarczyk:2021}.
It is this mapping that our agents explored and learned in this study.

In the future, we plan to extend our work to other problems, which do not necessarily need to be based on spatial navigation, such as when following a construction plan for building furniture from individual parts.
Moreover, affordances should be directly based on objects, besides local circumstances, such as a kettle for boiling water for tea. 
With regards to developmental psychology, further explorations with respect to curriculum generation \cite{Smith2018} might be interesting:
How do children generate their curriculum in comparison to our approaches?
How do they decide which experiences should never be forgotten, such as touching a hotplate?
The way we replace sequences in the training set can certainly be improved.
The challenge is to create a good set of training data and at the same time avoid forgetting of important experiences.
Even though we did not observe catastrophic forgetting in the above experiments, it would be interesting to examine whether the learning process exhibits self-stabilizing behavior if disturbed by impactful experiences.
In all of these cases, our study suggests that it will be advantageous to have local information available and to actively explore objects, entities, and locations in an epistemically-driven, active manner.

\section*{Acknowledgements}

This research was funded by the German Research Foundation (DFG) under Germany’s Excellence Strategy, EXC number 2064/1, project number 390727645, as well as within priority program SPP 2134, project ``DeepSelf: Emergence of Event-Predictive Agency in Robots'' (BU1335/14-1).
The authors thank the International Max Planck Research School for Intelligent Systems (IMPRS-IS) for supporting Fedor Scholz.

\bibliographystyle{unsrtnat}
\bibliography{neurips_2023}

\begin{thebibliography}{24}
\providecommand{\natexlab}[1]{#1}
\providecommand{\url}[1]{\texttt{#1}}
\expandafter\ifx\csname urlstyle\endcsname\relax
  \providecommand{\doi}[1]{doi: #1}\else
  \providecommand{\doi}{doi: \begingroup \urlstyle{rm}\Url}\fi

\bibitem[Butz and Kutter(2016)]{Butz2016}
Martin~V Butz and Esther~F Kutter.
\newblock \emph{How the mind comes into being: Introducing cognitive science
  from a functional and computational perspective}.
\newblock Oxford University Press, 2016.

\bibitem[Kuperberg(2021)]{Kuperberg2021}
Gina~R Kuperberg.
\newblock Tea with milk? a hierarchical generative framework of sequential
  event comprehension.
\newblock \emph{Topics in Cognitive Science}, 13\penalty0 (1):\penalty0
  256--298, 2021.

\bibitem[Gibson(1986)]{Gibson1986}
James~Jerome Gibson.
\newblock \emph{The ecological approach to visual perception}, volume~1.
\newblock Psychology Press New York, 1986.

\bibitem[Smith et~al.(2018)Smith, Jayaraman, Clerkin, and Yu]{Smith2018}
Linda~B Smith, Swapnaa Jayaraman, Elizabeth Clerkin, and Chen Yu.
\newblock The developing infant creates a curriculum for statistical learning.
\newblock \emph{Trends in cognitive sciences}, 22\penalty0 (4):\penalty0
  325--336, 2018.

\bibitem[Bottini and Doeller(2020)]{Bottini:2020}
Roberto Bottini and Christian~F. Doeller.
\newblock Knowledge across reference frames: Cognitive maps and image spaces.
\newblock \emph{Trends in Cognitive Sciences}, 24\penalty0 (8):\penalty0
  606--619, 2020.
\newblock ISSN 1364-6613.
\newblock \doi{10.1016/j.tics.2020.05.008}.

\bibitem[Buckner and Carroll(2007)]{Buckner:2007}
Randy~L. Buckner and Daniel~C. Carroll.
\newblock Self-projection and the brain.
\newblock \emph{Trends in Cognitive Sciences}, 11:\penalty0 49--57, 2007.

\bibitem[Tolman(1948)]{Tolman:1948}
Edward~C. Tolman.
\newblock Cognitive maps in rats and men.
\newblock \emph{Psychological Review}, 55:\penalty0 189--208, 1948.

\bibitem[O'Keefe and Nadel(1978)]{O'Keefe:1978}
J.~O'Keefe and L.~Nadel.
\newblock \emph{The Hippocampus as a Cognitive Map}.
\newblock Clarendon Press, Oxford, UK, 1978.

\bibitem[Ha and Schmidhuber(2018)]{Ha2018a}
David Ha and Jürgen Schmidhuber.
\newblock World models.
\newblock \emph{arXiv preprint arXiv:1803.10122}, Mar 2018.
\newblock \doi{10.5281/zenodo.1207631}.

\bibitem[Qi et~al.(2020)Qi, Mullapudi, Gupta, and Ramanan]{Qi2020}
William Qi, Ravi~Teja Mullapudi, Saurabh Gupta, and Deva Ramanan.
\newblock Learning to move with affordance maps.
\newblock \emph{arXiv preprint arXiv:2001.02364}, 2020.

\bibitem[Epstein et~al.(2015)Epstein, Aroor, Evanusa, Sklar, and
  Parsons]{Epstein2015}
Susan~L Epstein, Anoop Aroor, Matthew Evanusa, Elizabeth~I Sklar, and Simon
  Parsons.
\newblock Learning spatial models for navigation.
\newblock In \emph{Spatial Information Theory: 12th International Conference,
  COSIT 2015, Santa Fe, NM, USA, October 12-16, 2015, Proceedings 12}, pages
  403--425. Springer, 2015.

\bibitem[Scholz et~al.(2022)Scholz, Gumbsch, Otte, and Butz]{Scholz2022}
Fedor Scholz, Christian Gumbsch, Sebastian Otte, and Martin~V. Butz.
\newblock Inference of affordances and active motor control in simulated
  agents.
\newblock \emph{Frontiers in Neurorobotics}, 2022.
\newblock \doi{10.3389/fnbot.2022.881673}.

\bibitem[Rubinstein(1999)]{Rubinstein1999}
Reuven Rubinstein.
\newblock The cross-entropy method for combinatorial and continuous
  optimization.
\newblock \emph{Methodology and computing in applied probability}, 1\penalty0
  (2):\penalty0 127--190, 1999.

\bibitem[Der~Kiureghian and Ditlevsen(2009)]{Kiureghian2009}
Armen Der~Kiureghian and Ove Ditlevsen.
\newblock Aleatory or epistemic? does it matter?
\newblock \emph{Structural safety}, 31\penalty0 (2):\penalty0 105--112, 2009.

\bibitem[Hüllermeier and Waegeman(2021)]{Hüllermeier2021}
Eyke Hüllermeier and Willem Waegeman.
\newblock Aleatoric and epistemic uncertainty in machine learning: an
  introduction to concepts and methods.
\newblock \emph{Machine Learning}, 110\penalty0 (3):\penalty0 457--506, 2021.
\newblock ISSN 1573-0565.
\newblock \doi{10.1007/s10994-021-05946-3}.

\bibitem[Vlastelica et~al.(2021)Vlastelica, Blaes, Pinneri, and
  Martius]{Vlastelica2021}
Marin Vlastelica, Sebastian Blaes, Cristina Pinneri, and Georg Martius.
\newblock Risk-averse zero-order trajectory optimization.
\newblock In \emph{5th Annual Conference on Robot Learning}, 2021.

\bibitem[Kingma and Ba(2014)]{Kingma2014}
Diederik~P. Kingma and Jimmy Ba.
\newblock Adam: A method for stochastic optimization.
\newblock \emph{arXiv preprint arXiv:1412.6980}, 2014.

\bibitem[Briët and Harremoës(2009)]{briet2008}
Jop Briët and Peter Harremoës.
\newblock Properties of classical and quantum jensen-shannon divergence.
\newblock \emph{Physical Review A}, 79\penalty0 (5):\penalty0 052311, 2009.
\newblock \doi{10.1103/PhysRevA.79.052311}.

\bibitem[G{\"a}rdenfors(2014)]{Gaerdenfors:2014}
Peter G{\"a}rdenfors.
\newblock \emph{The Geometry of Meaning: Semantics Based on Conceptual Spaces}.
\newblock MIT Press, Cambridge, MA, 2014.

\bibitem[Mallot et~al.(2014)Mallot, Roehrich, and Hardiess]{Mallot:2014}
Hanspeter~A. Mallot, Wolfgang~G. Roehrich, and Gregor Hardiess.
\newblock A view-based account of spatial working and long-term memories: Model
  and predictions.
\newblock \emph{Proceedings of the 36th Annual Conference of the Cognitive
  Science Society}, pages 1911--1916, 2014.
\newblock Austin, TX: Cognitive Science Society.

\bibitem[Röhrich et~al.(2014)Röhrich, Hardiess, and Mallot]{Roehrich:2014}
Wolfgang~G. Röhrich, Gregor Hardiess, and Hanspeter~A. Mallot.
\newblock View-based organization and interplay of spatial working and
  long-term memories.
\newblock \emph{PLOS ONE}, 9\penalty0 (11):\penalty0 1--12, 11 2014.
\newblock \doi{10.1371/journal.pone.0112793}.

\bibitem[Buckner et~al.(2008)Buckner, Andrews-Hanna, and
  Schacter]{Buckner:2008}
Randy~L. Buckner, Jessica~R. Andrews-Hanna, and Daniel~L. Schacter.
\newblock The brain's default network.
\newblock \emph{Annals of the New York Academy of Sciences}, 1124\penalty0
  (1):\penalty0 1--38, 2008.
\newblock ISSN 1749-6632.
\newblock \doi{10.1196/annals.1440.011}.

\bibitem[Zacks(2020)]{Zacks:2020}
Jeffrey~M. Zacks.
\newblock Event perception and memory.
\newblock \emph{Annual Review of Psychology}, 71\penalty0 (1):\penalty0
  165--191, 2020.
\newblock \doi{10.1146/annurev-psych-010419-051101}.

\bibitem[Stawarczyk et~al.(2021)Stawarczyk, Bezdek, and Zacks]{Stawarczyk:2021}
David Stawarczyk, Matthew~A. Bezdek, and Jeffrey~M. Zacks.
\newblock Event representations and predictive processing: The role of the
  midline default network core.
\newblock \emph{Topics in Cognitive Science}, 13:\penalty0 164--186, 2021.
\newblock \doi{10.1111/tops.12450}.

\end{thebibliography}

\end{document}